\documentclass[conference]{IEEEtran}
\IEEEoverridecommandlockouts
\usepackage{cite}
\usepackage{booktabs}
\usepackage{amsmath,amssymb,amsfonts}
\usepackage{algorithmic}
\usepackage{graphicx}
\usepackage{textcomp}
\usepackage{subcaption}
\usepackage{tikz}
\usetikzlibrary{positioning,arrows.meta,calc}

\usepackage{caption}
\usepackage{xcolor}
\def\BibTeX{{\rm B\kern-.05em{\sc i\kern-.025em b}\kern-.08em
    T\kern-.1667em\lower.7ex\hbox{E}\kern-.125emX}}

\tikzstyle{startstop} = [rectangle, rounded corners, minimum width=3cm, minimum height=1.5cm,text centered, draw=black, fill=red!30]

\tikzstyle{process} = [rectangle, minimum width=3.5cm, minimum height=1cm, text centered, draw=black, fill=blue!30]
\tikzstyle{arrow} = [thick,->,>=stealth]

\begin{document}

\title{Real-Scale Island Area and Coastline Estimation using Only its Place Name or Coordinates}

\author{
    \IEEEauthorblockN{
        Quanyun Wu\IEEEauthorrefmark{1}, 
        Kyle Gao\IEEEauthorrefmark{1}, 
        Wentao Sun\IEEEauthorrefmark{1}, 
        Hongjie He\IEEEauthorrefmark{2}, 
        Yuhao Chen\IEEEauthorrefmark{1}, 
        David A. Clausi\IEEEauthorrefmark{1}, 
        Jonathan Li\IEEEauthorrefmark{1,2}
    }
    \IEEEauthorblockA{\IEEEauthorrefmark{1}University of Waterloo, Waterloo, ON, Canada\\ 
    \{q34wu, y56gao, w27sun, hongjie.he, yuhaochen1, dclausi, junli\}@uwaterloo.ca}
    \IEEEauthorblockA{\IEEEauthorrefmark{2}East China Normal University, Shanghai, China\\}
}

\maketitle
\begin{abstract}
Accurate measurement of island area and coastline length is crucial for coastal zone monitoring and oceanographic analysis. However, traditional measurement and mapping methods usually rely heavily on orthophotos, expensive airborne depth sensors, or dense ground control points, which face serious limitations of high labor costs, time-consuming efforts, and low operational efficiency in vast and inaccessible open sea environments. To overcome these challenges and break away from the reliance on manual field exploration, this paper proposes a geometrically consistent, real-scale island measurement framework based on pure monocular vision.  This project significantly reduces the mapping cost through a fully automated process and achieves high-efficiency measurement without prior GIS data. In our system pipeline, only the geographical coordinates or names of the target area need to be input to obtain a low-altitude surrounding image sequence. After obtaining the point clouds, a lightweight trajectory alignment algorithm (Umeyama) is used to restore the global physical scale, and the scaled model is orthorectified, enabling high-precision area and perimeter extraction directly on the 2D rasterized plane. We have fully verified this pipeline on four islands with different terrain features (covering natural landform islands and islands with complex artificial facilities). The experimental results show that the final measurement error of the system is stable at around 10\%, demonstrating excellent accuracy and robustness. Moreover, this framework has outstanding inference speed, requiring only 70 ms to process a single high-resolution image and generate point clouds, providing a highly practical new paradigm for large-scale marine and coastline mapping.
\end{abstract}

\begin{figure}[t]
    \centering
    \includegraphics[width=0.6\linewidth]{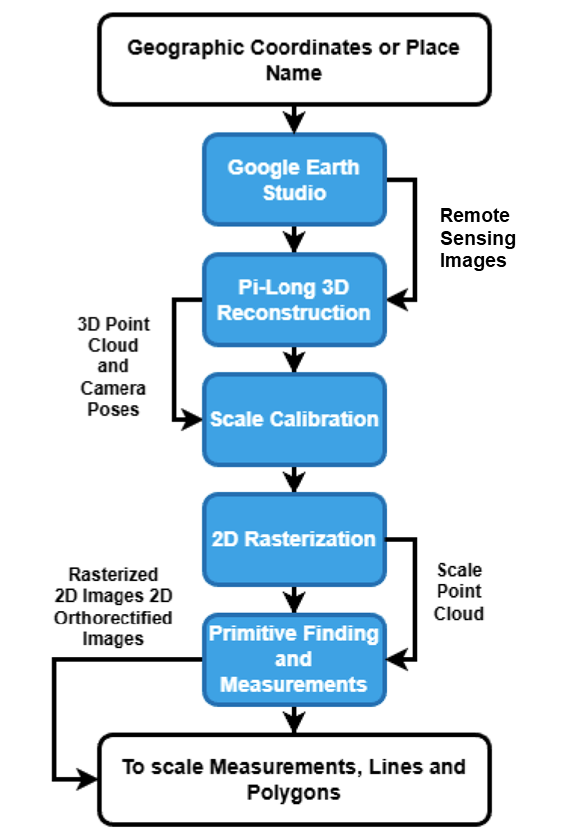}
    \caption{Flow chart of our pipeline. Starting from a place's geographic coordinates or identifying name and by leveraging Google Earth Studio, our pipeline creates a scale-calibrated 3D point cloud and allows for scale-accurate measurements of polygons and lines.}
\label{fig:pipeline}
\end{figure}



\section{Introduction}
Accurate measurement of island area and coastline length plays an irreplaceable fundamental role in coastal zone monitoring, marine navigation, and ecological analysis. However, traditional remote sensing and mapping methods often rely heavily on expensive airborne depth sensors (such as LiDAR) or time-consuming in-situ RTK GPS surveying \cite{goncalves2015unmanned}. In vast, inaccessible open-sea areas, these methods face serious limitations. To address this issue and break away from the reliance on manual on-site exploration, an urgent need arises for a low-cost, high-efficiency, and fast-response automated scale measurement solution.

To solve this problem, this paper proposes a fast, geometrically consistent real-scale island measurement system based entirely on monocular vision and web-GIS. Our approach builds on recent advances in vision-based 3D reconstruction and spatial understanding, which show strong scalability from indoor digital twins for embodied AI \cite{kitchentwin2026} to large-scale geographic mapping. The overall framework is shown in the system flowchart of Figure 1. This project significantly reduces the time and economic costs of mapping through highly automated pipelines, and proves that high-precision measurements can be completed without prior remote sensing or on-site data by relying on webGIS systems such as Google Earth Engine (GES) and only using the name or geographic coordinates of the target area. 

Our proposed pipeline directly takes the geographic coordinates of the target island to acquire low-altitude monocular image sequences via Google Earth Studio. To process these sequences, we first employ an advanced Transformer network (Pi-Long) to densely reconstruct a large-scale 3D point cloud and estimate camera poses~\cite{wang2025pi, deng2025vggt, pilongcode2025}. Rather than relying on costly ground control points, we resolve the inherent scale ambiguity of monocular vision by introducing a trajectory alignment algorithm (Umeyama~\cite{umeyama1991least}). Finally, to tackle severe occlusion and sea-surface reflections at oblique viewpoints, we propose a multi-view 2D-to-3D semantic back-projection strategy using SAM3~\cite{kirillov2023segment, carion2025sam}, which robustly isolates the island's geometry for high-precision, grid-based area measurement.

We evaluated the framework on four islands with diverse landscapes and complex facilities, comparing estimated areas to official data: Liberty Island ($56{,}685.68~\mathrm{m^2}$ vs. $59{,}560~\mathrm{m^2}$), Governors Island ($732{,}018~\mathrm{m^2}$ vs. $\sim 696{,}000~\mathrm{m^2}$), Somes Island ($212{,}699.2~\mathrm{m^2}$ vs. $\sim 249{,}000~\mathrm{m^2}$), and Ellis Island ($0.142~\mathrm{km^2}$ vs. $0.111~\mathrm{km^2}$). Despite oblique viewing angles, the system maintained $\sim 10\%$ average error and processed high-resolution ($1920\times1080$) images in 70~ms per frame, generating point clouds and measurements efficiently for large-scale marine studies.
\begin{figure*}[htbp]
    \centering
    \includegraphics[width=1\linewidth]{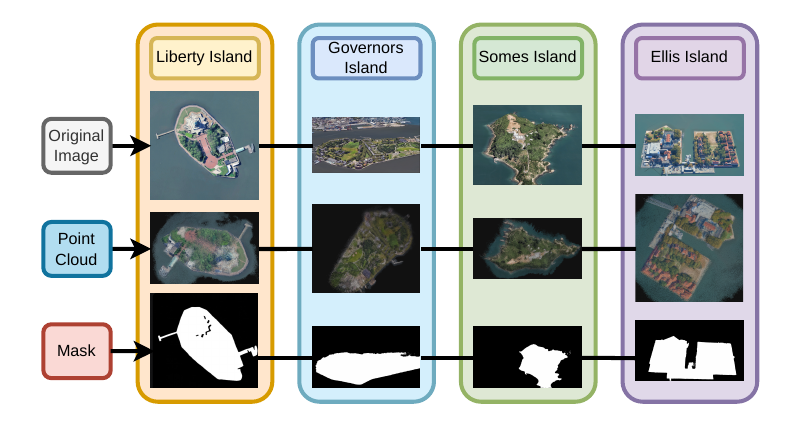}
    \caption{Qualitative results across four island datasets. (Row 1 to 4: Statue of Liberty Island, Governors Island, Somes Island, Ellis Island). Column 1: Original monocular remote sensing images. Column 2: Reconstructed real-scale colored 3D point clouds. Column 3: Top-down visualizations of the aggregated 3D-to-2D semantic masks and grid-based footprints used for area estimation.}
    \label{fig:pipeline}
\end{figure*}
\section{Method}
The framework proposed in this paper integrates monocular camera 3D reconstruction, scale calibration based on trajectory alignment, multi-view semantic back-projection, and metric calculation based on grids in sequence. The detailed descriptions of the sub-modules are as follows:

\textbf{Data Acquisition:} Monocular video acquisition from Google Earth Studio allows users to generate high-quality aerial sequences using only a place name or geographic coordinates. The platform’s virtual camera captures continuous views with controlled motion, providing synthetic image sequences and precise camera parameters suitable for 3D reconstruction and remote sensing analysis. The virtual camera poses can be easily extracted and stored.

\textbf{Global-scale restoration based on trajectory alignment:} Due to the inherent scale ambiguity of the monocular vision system, the reconstructed 3D space lacks the true physical proportions~\cite{campos2021orb}. While traditional Structure-from-Motion (SfM) pipelines like COLMAP \cite{schonberger2016structure} can reconstruct accurate geometry, they are computationally expensive for large-scale maritime scenes. Meanwhile, recent implicit representations like Neural Radiance Fields (NeRF) \cite{mildenhall2020nerf} excel at novel view synthesis but lack the explicit geometric boundaries required for rapid metric footprint extraction. To avoid relying on ground control points (GCPs), we align the reconstructed camera trajectory with the real reference trajectory in the GES virtual camera data to restore the geometric scale. Let $x_{i} \in \mathbb{R}^{3}$ be the translation vector (not scaled) of the reconstructed camera pose, and $y_{i} \in \mathbb{R}^{3}$ be the corresponding real camera center. We estimate a global similarity transformation $Sim(3)$ by minimizing the alignment error, which consists of a scaling factor $c$, a rotation matrix $R \in SO(3)$, and a translation vector $t \in \mathbb{R}^{3}$: $$\min_{c,R,t} \frac{1}{N} \sum_{i=1}^{N} ||y_{i} - (cRx_{i} + t)||^{2}$$ The above minimization problem is solved by the Umeyama algorithm to obtain a closed-form solution. The calculated scaling factor $c$, rotation $R$, and translation $t$ are combined into a homogeneous transformation matrix $H \in \mathbb{R}^{4 \times 4}$. To apply this transformation to the camera pose without disrupting the orthogonality of the camera rotation matrix, we extract the pure rotation component $R_{align} = H[:3, :3] / c$ from $H$. The update rule for the camera pose is as follows: $$R_{new} = R_{align} R_{orig}, \quad t_{new} = (cR)t_{orig} + t$$ Finally, the global similarity transformation $Sim(3)$ is applied to the entire reconstructed 3D point cloud to accurately restore its physical size in the real world.

\textbf{Multi-view Semantic Segmentation Based on the Basic Model:} Although the point cloud scale was restored, sea surface noise and occlusion issues still exist at tilted viewpoints. To precisely extract the target islands from the surrounding ocean and background noise, we introduce a multi-view 2D-to-3D projection strategy that combines the Segment Anything Model 3 (SAM3). For each valid image frame $i$, SAM3 generates a high-precision 2D binary mask $M_{i}$. Given a real-scale point cloud with $N$ points, we use the inverse matrix of the real-scale camera extrinsic parameters $T_{c2w}$ to transform the world coordinate system point $P_{w} = [X_w, Y_w, Z_w, 1]^{T}$ to the local camera coordinate system: $$P_{c} = R_{w2c}P_{w} + t_{w2c}$$ To ensure geometric validity and reduce computational redundancy, the system filters out the points located behind the camera plane (i.e., $Z_{c} \le 0.1$). The remaining valid 3D points will be orthographically projected onto the 2D image plane using the camera intrinsic matrix $K$: $$u = \lfloor f_x \frac{X_c}{Z_c} + c_x \rfloor, \quad v = \lfloor f_y \frac{Y_c}{Z_c} + c_y \rfloor$$ where $(f_x, f_y)$ are the scaled focal lengths and $(c_x, c_y)$ are the principal point coordinates. For each point mapped to the valid pixel coordinates $(u, v)$, we query the corresponding 2D SAM3 mask $M_{i}$. If $M_{i}(u, v) > 0$, then this point is marked as an island point. To address the occlusion problem in tilted viewpoints and ensure the integrity of the geometric structure, we obtain the global union (Global Union) of the hit mask points across all viewpoints: $$M_{global} = \bigcup_{i=1}^{F} \{ P_{w} \mid M_{i}(u_i, v_i) > 0 \}$$ Through this multi-view fusion logic, the system fully utilizes the semantic understanding ability of SAM3 across viewpoints, iteratively constructing a robust and unrestricted by a single viewpoint complete 3D representation of the islands.
\begin{table*}[htbp]
    \centering
    \normalsize 
    \renewcommand{\arraystretch}{1.0} %
    \caption{QUANTITATIVE AREA ESTIMATION RESULTS ACROSS VARIOUS ISLANDS}
    \label{tab:area_estimation}
    \begin{tabular}{l|r|r|r}
        \toprule
        \textbf{Island (Target Island)} & \textbf{Ground Truth ($m^2$)} & \textbf{Estimation Result ($m^2$)} & \textbf{Relative Error (\%)} \\
        \midrule
        Statue of Liberty Island & 59,560.0 & 56,685.68 & -4.83\% \\
        Governors Island & 696,000.0 & 623,323.31 & -10.44\% \\
        Somes Island & 249,000.0 & 280,875.98 & +12.80\% \\
        Ellis Island & 111,000.0 & 119,120.51 & +7.32\% \\
        \bottomrule
    \end{tabular}
\end{table*}
\textbf{Grid-based area estimation and height map generation:} Utilizing modern 3D data processing pipelines \cite{zhou2018open3d}, we extract the pure 3D island point cloud, discard the Z axis coordinate, and project the points onto a 2D plane to obtain $P_{2D} = (X, Y)$. To accurately calculate the projected area of the ground surface and avoid being affected by the uneven local density of the point cloud, we propose a discrete grid rasterization (Grid-based Rasterization) method. To ensure scale adaptability, the physical space is quantized into a regular square grid with a dynamic resolution $s$. The system first calculates the average nearest neighbor spacing $dist_{avg}$ of the local point cloud and defines the optimal grid size as $s = \max(2 \times dist_{avg}, 0.05)$ meters, thereby avoiding the discretization error caused by a fixed grid. The two-dimensional coordinates are mapped to discrete grid indices: $$G_{x,y} = \lfloor \frac{P_{2D}}{s} \rfloor$$ Subsequently, the system extracts all the occupied independent grid sets $U_{cells}$. The total physical area $A$ can be calculated analytically by multiplying the number of independent grids $|U_{cells}|$ with the area of a single grid $s^{2}$: $$A = |U_{cells}| \times s^{2}$$ At the same time, the system maps the occupied grids to a pixel array to generate a top-down orthogonal visualization image. Additionally, using the previously retained $Z$ axis coordinate (i.e., physical height), the system normalizes and maps it to a grayscale intensity value, directly generating a black-and-white height map (white pixels represent higher altitudes). This process completely avoids the complexity of traditional polygon contour fitting and provides a high-precision measurement scheme for irregular coastlines.

\section{Results and Discussion}
\textbf{Experimental Setup and Dataset:} We evaluate our framework on a newly constructed dataset comprising four distinct islands on four islands with different terrain features and complexity of man-made structures: Liberty Island (dominated by regular man-made edges), Governor Island (extensive in area and featuring complex landforms), Somes Island (a lush natural island), and Ellis Island (containing numerous historical buildings and connected bridges). All remote sensing images were obtained through Google Earth Studio and had a resolution of $1920 \times 1080$. The experiments were run on a workstation equipped with a single NVIDIA RTX 4090 GPU.

\textbf{Qualitative Evaluation:} Figure \ref{fig:pipeline} presents the multi-modal qualitative visualization results of this system on four islands. The figure demonstrates that the Pi-Long scale restoration pipeline achieves robust global geometry recovery across diverse island scenes, despite the absence of nadir viewpoints and the presence of strong perspective distortions. The reconstructed colored point clouds preserve consistent scale and structural integrity, indicating that the method effectively mitigates scale ambiguity and drift typically observed in monocular or sparse-view reconstruction settings. Furthermore, the alignment between the reconstructed 3D geometry and the rasterized occupancy maps suggests that the multi-view back-projection process produces geometrically coherent and semantically consistent island footprints. This consistency implies that the system not only captures accurate large-scale structure but also enables reliable downstream spatial analysis, such as area estimation and boundary extraction, under challenging real-world imaging conditions.



\textbf{Quantitative Evaluation:} To quantitatively evaluate the accuracy of the measurement, we compared the estimated area calculated by the system with the officially published true area. It is worth noting that to eliminate the influence of uneven point cloud density distribution on the rasterized area, this system introduced an adaptive grid resolution mechanism. The optimal grid size ($s = \max(2 \times dist_{avg}, 0.05)$ meters) was automatically derived by calculating the average nearest neighbor distance of the point cloud. The experiment used relative error (RE) as the core evaluation metric. The updated multi-view segmentation and measurement results are shown in Table \ref{tab:area_estimation}. As shown in Table \ref{tab:area_estimation}, this system demonstrated extremely high measurement accuracy on the Statue of Liberty Island and Ellis Island. Especially for Ellis Island, which contains a large number of complex historical buildings and connected bridges, thanks to the multi-view mask voting mechanism in the new pipeline (Min Votes = 3) and Z-Buffer occlusion elimination (Depth Tolerance = 2.0m), the system successfully filtered out the surrounding free piers and water surface reflection noise points, strictly controlling the error within an excellent level of +7.32\%. For Governors Island, the area was under an approximately 10.44\% reasonable underestimate. This is mainly attributed to the presence of a large area of gentle intertidal zone and shallow water area in the southern part of the island. Under the strict intersection voting of multiple views, the edge water areas were not included in the hard land area. In contrast, Somes Island showed a +12.80\% positive error. The main reason for this is that this island is a nature reserve, with extremely dense vegetation (Canopy) on the edge and a complex reef group along the coastline. Due to the SAM3 model's understanding of "island" during multi-view semantic segmentation, it tends to include all continuous terrestrial vegetation canopies and nearshore exposed reefs, resulting in the extracted three-dimensional point cloud contour from the inverse projection exceeding the officially defined pure surface coastline benchmark. Overall, the average absolute error of these four islands with high topographic differences was stable at around 8.84\%. Under the strict conditions of no need for any ground control points (GCPs), no nadir views, and relying solely on monocular oblique photography, the system proved its extremely high robustness in high-precision marine surveying through adaptive grid rasterization and multi-view geometric fusion.

\section{Conclusion}
This paper presents a fast and robust framework for real-scale island measurement using pure monocular vision. Our system eliminates the reliance on expensive airborne sensors or dense ground control points. We achieve this by integrating web-GIS-based remote sensing monocular video rendering with advanced Transformer-based 3D reconstruction and a trajectory alignment algorithm. This combination effectively resolves scale ambiguity inherent in monocular sequences. The proposed multi-view semantic back-projection strategy further ensures geometric consistency by filtering sea-surface noise and occlusions. Preliminary experimental evaluations across diverse islands demonstrate a stable measurement error of approximately 10\%. High inference speeds of 70 ms per image highlight the practical utility of our pipeline. This method provides an efficient new paradigm for large-scale maritime mapping and coastal zone monitoring. It proves that high-precision spatial analysis is possible using only geographic coordinates and monocular imagery. Ongoing work for a future journal publication expands these tests to complex reef environments and further optimizes the framework design for real-time edge deployment.

\bibliographystyle{IEEEtran}
\bibliography{ref}

\end{document}